\documentclass{IOS-Book-Article}

\usepackage[T1]{fontenc}
\usepackage{mathptmx}
\usepackage{soul}\setuldepth{article}

\usepackage{xspace}
\usepackage{graphicx}
\usepackage{array}
\hypersetup{pageanchor=false,
  pdftitle={ABDA-NL: A Natural-Language Scenario Explorer for Argument-Based Reasoning},
  pdfauthor={Shawn Bowers, Martin Caminada, Haoyang Liu, Bertram Lud\"ascher}}

\AtBeginDocument{}
\newcommand{\abdanl}{\textsc{abda-nl}\xspace}
\newcommand{\abda}{\textsc{abda}\xspace}
\newcommand{\aspicm}{\textsc{aspic-}\xspace}
\newcommand{\aspicp}{\textsc{aspic+}\xspace}
\newcommand{\figref}[1]{Fig.\,\ref{#1}}
\newcommand{\secref}[1]{Section~\ref{#1}}
\newcommand{\ui}[1]{\emph{#1}}
\newcommand{\kw}[1]{\textsc{#1}}
\newcommand{\move}[1]{\textsc{#1}}

\begin{document}

\pagestyle{headings}
\def\thepage{}
\begin{frontmatter}

\title{\abdanl: A Natural-Language Scenario Explorer for
    Argument-Based Reasoning}
\subtitle{Extended Version of the COMMA 2026 Demonstration Abstract}

\markboth{}{}

\author[A]{\fnms{Shawn} \snm{Bowers}},
\author[B]{\fnms{Martin} \snm{Caminada}},
\author[C]{\fnms{Haoyang} \snm{Liu}},
\author[C]{\fnms{Bertram} \snm{Lud\"ascher}}

\runningauthor{Bowers et al.}
\address[A]{Gonzaga University, Washington, USA}
\address[B]{Cardiff University, Wales, UK}
\address[C]{University of Illinois Urbana-Champaign, USA}

\begin{abstract}
  \abdanl adds a natural-language interface to \abda, a system for
  argument-based discussion using \aspicm knowledge bases under
  grounded semantics.  Users see which conclusions are accepted,
  rejected, or undecided, open an interactive rendering of the
  grounded discussion game to learn why, explore what-if alternatives
  by suspending assumptions and rules or changing preferences, ask
  questions that are answered from a scenario's reference documents,
  and author new facts, assumptions, and rules in plain
  English.  A large language model provides the bridge between
  language and formalism: it answers questions from the documents and
  the current state of the scenario, and it translates plain-English
  edits into candidate formal statements.  The deterministic \abda
  engine remains the sole source of arguments, attacks, and
  acceptance labels, and every proposal of the model is validated and
  confirmed by the user before it takes effect.
\end{abstract}

\end{frontmatter}

\markboth{}{}

\section{Introduction}

\begin{figure}[t]
  \centering
  \includegraphics[width=\textwidth]{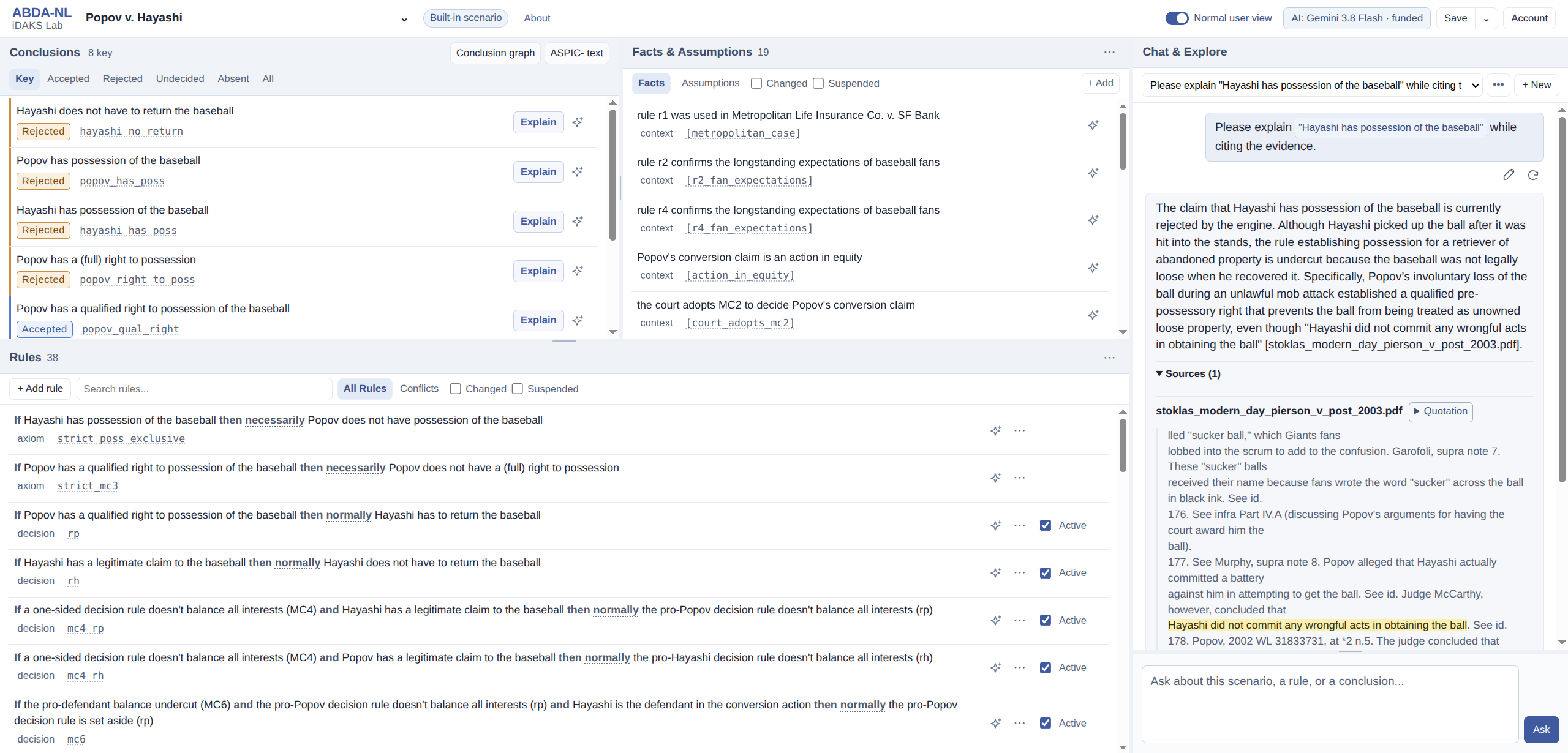}
  \caption{\abdanl on \emph{Popov v.\ Hayashi} \cite{prakken}: the
    \aspicm scenario explorer with corpus-grounded chat.}
  \label{fig:screenshot}
\end{figure}

Informal argumentation typically involves discussion in which parties
exchange claims, raise objections, and defend their positions
\cite{WK95}.  In contrast, formal argumentation often focuses on
computing which claims are accepted from a graph of arguments and
their attacks, treating acceptance as a static property rather than
the outcome of dialog \cite{GDG}.  The \abda (Argument-Based
Discussion using \aspicm) system \cite{CU20} allows users to play a
grounded discussion game over \aspicm knowledge bases \cite{CMO14}, a formalism
with strict and defeasible rules, strong negation, defeasible
assumptions, and rule preferences.  However, \abda still operates at
the formal level, where propositions appear as \aspicm symbols,
discussion games are played via formal move primitives, and
explanations of moves and outcomes are not given.  We present \abdanl
({\small\href{https://github.com/idaks/ABDA-NL}{\texttt{github.com/idaks/ABDA-NL}}}),
a web-based tool that extends \abda with a natural-language user
interface: users inspect, explore, and extend scenarios without
needing to engage in \aspicm notation, while \abda's reasoning over
the argumentation framework stays authoritative, so that conclusions
are formally grounded, inspectable, and contestable.
\figref{fig:screenshot} illustrates the tool on a well-known legal
case (\emph{Popov v.\ Hayashi}) formally modeled using strict and
defeasible rules in \cite{prakken}.

This paper extends the two-page abstract that accompanies our
demonstration at COMMA 2026 \cite{BCLL26} with the formal background
(\secref{sec:background}), a description of the system and a
walkthrough of a session (Sections \ref{sec:system} and
\ref{sec:using}), an account of how the language model and the argumentation engine
share the work (\secref{sec:llm}), and the features added since the
camera-ready version (\secref{sec:new}).

\section{Background: \aspicm and the Grounded Discussion Game}
\label{sec:background}

An \aspicm knowledge base \cite{CMO14} consists of \emph{strict
  rules} $p_1,\ldots,p_n \rightarrow c$, which admit no exceptions,
and \emph{defeasible rules} $p_1,\ldots,p_n \Rightarrow c$, which
hold normally but may have exceptions.  Facts are strict rules without
premises, and assumptions are defeasible rules without premises.
Literals are atoms or their strong negations (written $-p$), and each
defeasible rule has a name $r$ that can itself be negated, so that a
rule concluding $-r$ expresses an exception to $r$.  Arguments are
built by chaining rules.  An argument $A$ attacks an argument $B$ if
the conclusion of $A$ is the negation of the conclusion of $B$ or of
one of $B$'s sub-arguments (\emph{rebut}), or the negation of the
name of a defeasible rule used in $B$ (\emph{undercut}).  Because
facts and assumptions are rules without premises, an attack on a
premise, called undermining in \aspicp, is simply a rebut of that
sub-argument here.  \aspicm differs from \aspicp \cite{MP14} in allowing \emph{unrestricted}
rebut, that is, rebuttal of any sub-argument regardless of whether
its top rule is strict or defeasible, a choice that an empirical study found to be closer to how
people evaluate arguments \cite{YXL18}.  Preferences are expressed by assigning
defeasible rules and assumptions to strength levels, called
\emph{blocks} in \abda.  Arguments are compared by the weakest-link
principle with democratic ordering: a rebut succeeds only if the
attacker is not weaker than the sub-argument it attacks, whereas
undercuts always succeed.  The arguments and successful attacks form
a Dung-style argumentation framework \cite{Dun95}, which \aspicm
evaluates under grounded semantics.  Every argument receives one of
the labels \emph{in}, \emph{out}, or \emph{undec} \cite{GDG}, and a
literal is \emph{accepted} if it is the conclusion of an \emph{in}
argument, \emph{undecided} if it is the conclusion of an \emph{undec}
argument but of no \emph{in} argument, and \emph{rejected} otherwise.

Grounded semantics can be explained through the \emph{Grounded
  Discussion Game} \cite{Cam15a}, a dialog between a proponent and an
opponent that is sound and complete for grounded semantics: the
proponent has a winning strategy for an argument $A$ if and only if
$A$ is in the grounded extension.  The game has four moves.
\move{htb}$(A)$, ``$A$ \emph{has to be} the case,'' is the
proponent's move; it opens the game with the main argument and is
later used to defend against challenges.  \move{cb}$(B)$, ``$B$
\emph{can be} the case,'' is the opponent's move; $B$ must attack
the argument of the most recent \move{htb} move that has not yet been
conceded.  \move{concede}$(A)$ is played by the opponent once every
attacker of $A$ has been retracted, and \move{retract}$(B)$ once some
attacker of $B$ has been conceded.  The discussion ends when no move is possible, in particular once
an argument has been used twice, by one player or by both; the
proponent wins if the main argument has been conceded, and the
opponent wins otherwise.  \abda \cite{CU20}
lets a user play this game against the system through these move
primitives over formal \aspicm arguments; \abdanl renders the same
dialectical structure in natural language (\secref{sec:using}).

\section{\abdanl: Scenarios, Architecture, and Interface}
\label{sec:system}

\paragraph{Scenarios}
A \emph{scenario} in \abdanl consists of an \aspicm knowledge base, a
glossary, and a corpus of reference documents.  The knowledge base is
stored as a YAML file that declares facts, assumptions, intermediate
propositions, key conclusions, and rules, each with an identifier;
assumptions and defeasible rules also carry a strength block and an
\emph{active} flag that allows them to be suspended.  The glossary
maps every identifier to a plain-language description and, where
needed, to the wording of its negation.  The four rules below, taken from
\emph{Popov v.\ Hayashi} and shown with their rendering, illustrate
the format.  The first two rebut each other, since their conclusions
are opposite literals.  The fourth concludes the negation of the name
of the third, so an argument built with it undercuts every argument
that uses rule \texttt{wt1}.  Its first premise is an assumption: the scenario treats the
validity of each commonsense or legal principle as a defeasible
premise that can itself be contested.

\begin{center}\footnotesize
\begin{tabular}{@{}p{1.4em}@{}>{\raggedright\arraybackslash}p{0.86\textwidth}@{}}
\multicolumn{2}{@{}l@{}}{\texttt{popov\_qual\_right => -hayashi\_no\_return [rp]}}\\
& \kw{if} Popov has a qualified right to possession of the baseball
\kw{then normally} Hayashi has to return the baseball\\[2pt]
\multicolumn{2}{@{}l@{}}{\texttt{hayashi\_legit\_claim => hayashi\_no\_return [rh]}}\\
& \kw{if} Hayashi has a legitimate claim to the baseball \kw{then
  normally} Hayashi does not have to return the baseball\\[2pt]
\multicolumn{2}{@{}l@{}}{\texttt{popov\_testimony\_control => popov\_had\_full\_control [wt1]}}\\
& \kw{if} Popov testified that he had full control of the baseball
when he was attacked \kw{then normally} Popov had full control when
he was attacked\\[2pt]
\multicolumn{2}{@{}l@{}}{\texttt{cs3\_valid, testimony\_inconsistent => -wt1 [cs3]}}\\
& \kw{if} inconsistent testimony is unreliable (cs3) \kw{and} Popov's
testimony is inconsistent, ambiguous, and unconvincing \kw{then
  normally} Popov's control-testimony doesn't stand on its own (wt1)
\end{tabular}
\end{center}

\noindent
Each identifier is rendered by its description, a negated literal
by the wording of its negation, and the negated rule name
\texttt{-wt1} by the wording recorded for the exception to
\texttt{wt1}.  The knowledge base of \emph{Popov v.\ Hayashi} is based on Prakken's
reconstruction \cite{prakken}, and its corpus holds the court's
ruling, a law review article, a Wikipedia article, and a short blog
commentary.  The corpus serves only as context for the language
model; it never enters the reasoning.  \abdanl ships with six scenarios: \emph{Popov
  v.\ Hayashi} (19 facts, 20 assumptions, 38 rules, and 8 key
conclusions, yielding 77 arguments and 24 attacks) and five smaller
decisions from everyday life, medicine, environmental management,
and sports.

\paragraph{Architecture}
\abdanl has three layers.  The \emph{deterministic core} embeds the
\abda engine as an in-process library.  After every change it
rebuilds all arguments and attacks from the current knowledge base,
computes the grounded labeling, and derives the label of every
proposition.  This takes milliseconds for the built-in scenarios, and
limits on the number of arguments and attacks the engine will
construct keep a pathological knowledge base from stalling the
server.  The \emph{rendering layer} turns formal objects into
natural language purely through templates over the glossary: rules
become \kw{if}/\kw{then} sentences that read ``necessarily'' for
strict and ``normally'' for defeasible rules, and the panels, the
explanation trace, the conclusion graph, and the raw \aspicm listing
are all generated from the same computed framework.  The third layer
consists of optional \emph{language-model services} for question answering and
natural-language authoring (\secref{sec:llm}); without them, \abdanl
runs in a basic mode in which every argumentation feature remains
available.

\paragraph{Interface}
The main view (\figref{fig:screenshot}) has four panels: a
conclusions dashboard (upper left) with the labels \emph{accepted},
\emph{rejected}, and \emph{undecided}, plus \emph{absent} for a key
conclusion that the current knowledge base cannot derive either way;
the facts and assumptions (upper center); the rules (bottom); and a
chat panel (right).  Every item shows its identifier next to its
description, so that the natural-language and the formal views stay
connected, and filters restrict each list.

\section{Using the Demo}
\label{sec:using}

\paragraph{Inspect conclusions}
After loading a scenario, the user reads the key conclusions and
their labels.  In \emph{Popov v.\ Hayashi}, for example, ``Popov has
a legitimate claim to the baseball'' and ``Hayashi has a legitimate
claim to the baseball'' are both accepted, whereas ``Popov has
possession of the baseball'' is rejected.

\begin{figure}[t]
  \centering
  \fbox{\includegraphics[width=0.559\textwidth]{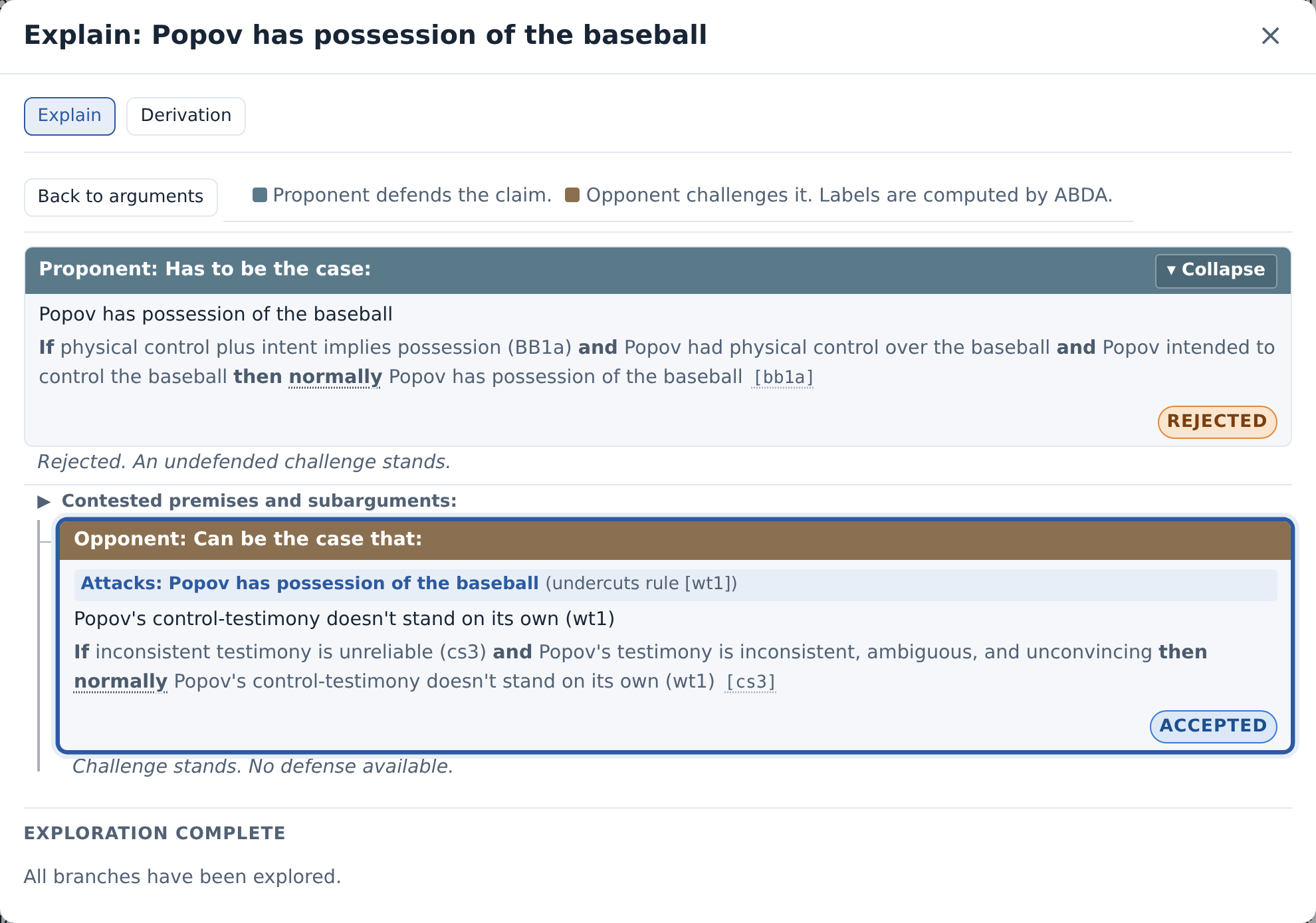}}
  ~
  \fbox{\includegraphics[width=0.371\textwidth]{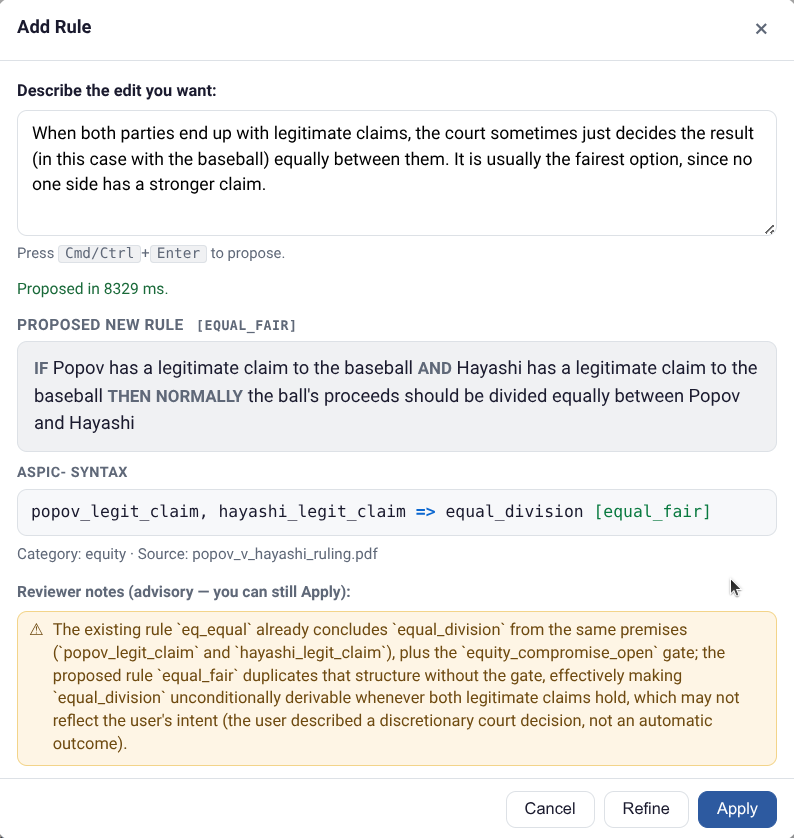}}
  \caption{Interactive explanation via the discussion game (left) and
    natural-language rule authoring (right).}
  \label{fig:game-and-rule}
\end{figure}

\paragraph{Explain}
Clicking \ui{Explain} on a conclusion opens the explanation view
(\figref{fig:game-and-rule}, left), which is organized as the
grounded discussion game would unfold.  The root card, labeled
\emph{Proponent: Has to be the case}, is the proponent's \move{htb}
move for an argument with the chosen conclusion; it shows the rule that derives
the conclusion and expands to its premises and sub-arguments down to
the facts.  Each card labeled \emph{Opponent: Can be the case that} is an
opponent's \move{cb} move: a counter-argument that rebuts the
conclusion of a sub-argument or undercuts one of its defeasible
rules, with the kind of attack stated.  Expanding a challenge reveals the proponent's replies.  Rather than taking turns, the
user explores the whole tree, and the end of each branch is worded as
the game outcome would be: a challenge is \emph{retracted} when a
defense exists, \emph{stands} when no defense is available, and a
claim is \emph{conceded} when every challenge against it has been
answered.  Derivations of the opposite conclusion that produce no attack, as
happens when a preference blocks a rebut, are listed where the branch
ends, and mutual attacks among undecided arguments are reported as a
cycle.  Only
counter-arguments that matter under grounded semantics are shown,
which keeps the tree small.

\paragraph{Explore what-if alternatives}
Assumptions and defeasible rules have a toggle that suspends them.
Before a toggle takes effect, \abdanl recomputes the scenario and
shows which conclusions would change label; activating the equity
assumption in \emph{Popov v.\ Hayashi}, for example, turns ``the ball's proceeds should be divided equally between Popov and
Hayashi'' from absent to undecided.
The user can then apply or cancel.  The \ui{Conflicts} tab of the rules panel lists pairs of defeasible
rules whose conclusions contradict each other (an assumption can
take the place of a rule) and lets the user prefer one side or
restore equal priority, which \abdanl translates into strength
blocks.  Conclusions whose label changed are highlighted, an
indicator counts the changes from the baseline, and \ui{Reset}
restores it, with an undo.

\paragraph{Ask questions}
The chat panel answers questions about the case from the corpus and
from the current state of the scenario, citing the documents it used
(\figref{fig:screenshot}, right).  A sparkle icon next to any
conclusion, fact, assumption, or rule adds that item to the question
being composed.

\paragraph{Author statements}
Through \ui{+~Add} in the facts panel, \ui{+~Add rule}, or
\ui{Modify with AI} on a rule, the user describes the intended
statement in plain English.  \abdanl proposes a formalization, shows it both as a
sentence and in \aspicm syntax together with advisory review notes
(\figref{fig:game-and-rule}, right), and lets the user refine the
request, cancel, or apply.  On apply, \abda recomputes all labels.

\paragraph{Inspect the framework}
\ui{Conclusion graph} draws the argumentation framework aggregated
by conclusion, with colors for labels and distinct edge styles for
mutual rebuts, one-way rebuts, and undercuts; any node opens the
derivations of its literal.  \ui{\aspicm text} lists the current
knowledge base, including glossary comments and suspension markers,
in a form that can be re-imported.

\section{The Role of the Language Model}
\label{sec:llm}

\abdanl pairs a symbolic engine with a large language model and gives
each the part of the work it is good at.  \abda decides what follows
from a knowledge base: it constructs the arguments and attacks and
computes the grounded labels, deterministically and for every change.
The language model handles language: it answers questions from the
scenario's documents and current state, and it turns a plain-English
request into a candidate formal statement.  The text of the interface
itself, from rule sentences to the explanation trace, is generated
from the glossary by templates, so a scenario can be explored,
explained, and edited by hand without any model, and whatever a model
contributes is visible as such.  The two calls, and the checks around each, are described below.

\paragraph{Grounded question answering}
For each question, \abdanl assembles context of two kinds.  From the
corpus it selects passages by a deterministic lexical ranking that
keeps at least one excerpt from every document and records where each
excerpt occurs.  From the engine it builds a description of the
current state: the label of every key conclusion together with the
rules that support, oppose, or undercut it, the state of every
assumption, the arguments with their labels and defeats, and any
changes made to the baseline.  An item added to the question through its sparkle icon is passed by
identity, so that its formal definition and its role in the framework
are in view.  The prompt names the engine as the authority
on what is warranted and asks the model to restate its labels, to
quote sources verbatim and cite the file each quote comes from, to
decline questions outside the scenario, and to ignore instructions
that appear inside documents.  Before an answer is shown, the server
checks every citation against the corpus and every quotation against
the exact passages it supplied; a draft that fails is returned to the
model once with the problem, and if the second draft fails too, the
user sees a short notice.  Verified
excerpts are displayed with the answer and can be opened at their
location in the source (\figref{fig:screenshot}, right).  Within
these bounds the model does what a symbolic engine cannot: it relates
the formal outcome to the facts of the case in fluent prose, grounds
its account in the documents, and answers follow-up questions in
context.

\paragraph{Natural-language authoring}
When a user describes a new fact, assumption, or rule in plain
English, the model acts as a translator into \aspicm.  It receives
the request together with the current state of the scenario and must
return a structured proposal through a schema-constrained tool call:
the operation, premises and conclusion as identifiers, the rule type
and strength block, and a glossary entry for every new identifier.  The prompt
directs it to reuse existing identifiers wherever the request refers
to an existing statement.  The proposal is then checked
deterministically.  Malformed payloads, missing or colliding
identifiers, references to unknown rules, and structurally impossible
combinations are rejected; the model is then told what was wrong and
asked for a new proposal, and if three attempts fail, the request is
abandoned and the problems are reported to the user.  A second, advisory model call
reviews the surviving proposal against the scenario and flags
concerns such as a mismatch between the request and the
formalization, a description that contradicts the formal content, or
redundancy with an existing rule; these notes are shown but do not
block.  Finally, the user sees the proposal both as a sentence and in
\aspicm syntax (\figref{fig:game-and-rule}, right) and decides
whether to refine, cancel, or apply it.  Nothing is applied
automatically, and an applied rule is treated exactly like a
hand-written one.

The formalization of a free-text request is not unique, and two runs
may propose different candidates, for
instance one that reuses an existing proposition and one that
introduces a new one.  The design does not depend on the model being
deterministic: whatever is applied is well-formed, refers only to
declared symbols, has been seen by the user in its exact formal
form, and has its consequences computed by the engine.  Scenario files and imported \aspicm text are parsed and validated
by a deterministic loader.

\paragraph{Models and their qualification}
The hosted service offers several models from different providers,
and users may also bring a key of their own.  A model is admitted to
this set only after passing an evaluation suite of question-answering
and authoring cases over all built-in scenarios, run repeatedly:
every case must pass in every repetition, every quotation must be
found verbatim in the corpus, every acceptance status reported must
agree with the engine, and every proposal must match the expected
operation.  If a model call fails, the chat says so and everything else continues to
work.

\section{New Features Since the Camera-Ready Version}
\label{sec:new}

\paragraph{Hosted service}
\abdanl now runs as a public service at
\href{https://demo.abda-nl.org}{\texttt{demo.abda-nl.org}}, so that anyone can explore the built-in scenarios without installing
anything or signing in.
Model-assisted features require a verified email sign-in; the user
then picks one of several qualified models and pays for it either
from funded trial credit, offered to a limited number of users, or
with an API key of their own, which is held in the browser tab only.
The interface is keyboard-operable and passes automated
accessibility checks at WCAG levels A and AA.

\begin{figure}[t]
  \centering
  \fbox{\includegraphics[width=0.373\textwidth]{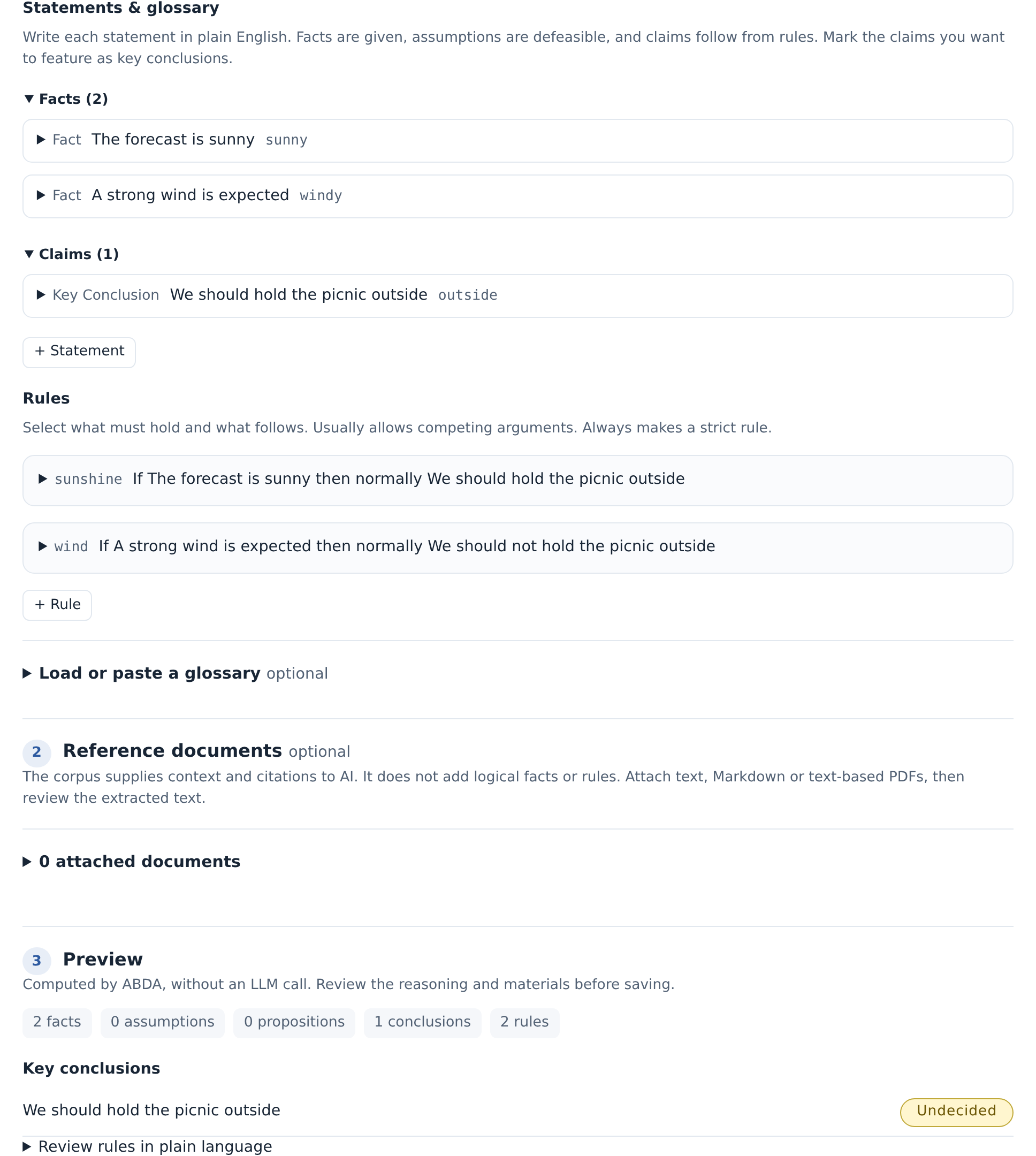}}
  ~
  \fbox{\includegraphics[width=0.557\textwidth]{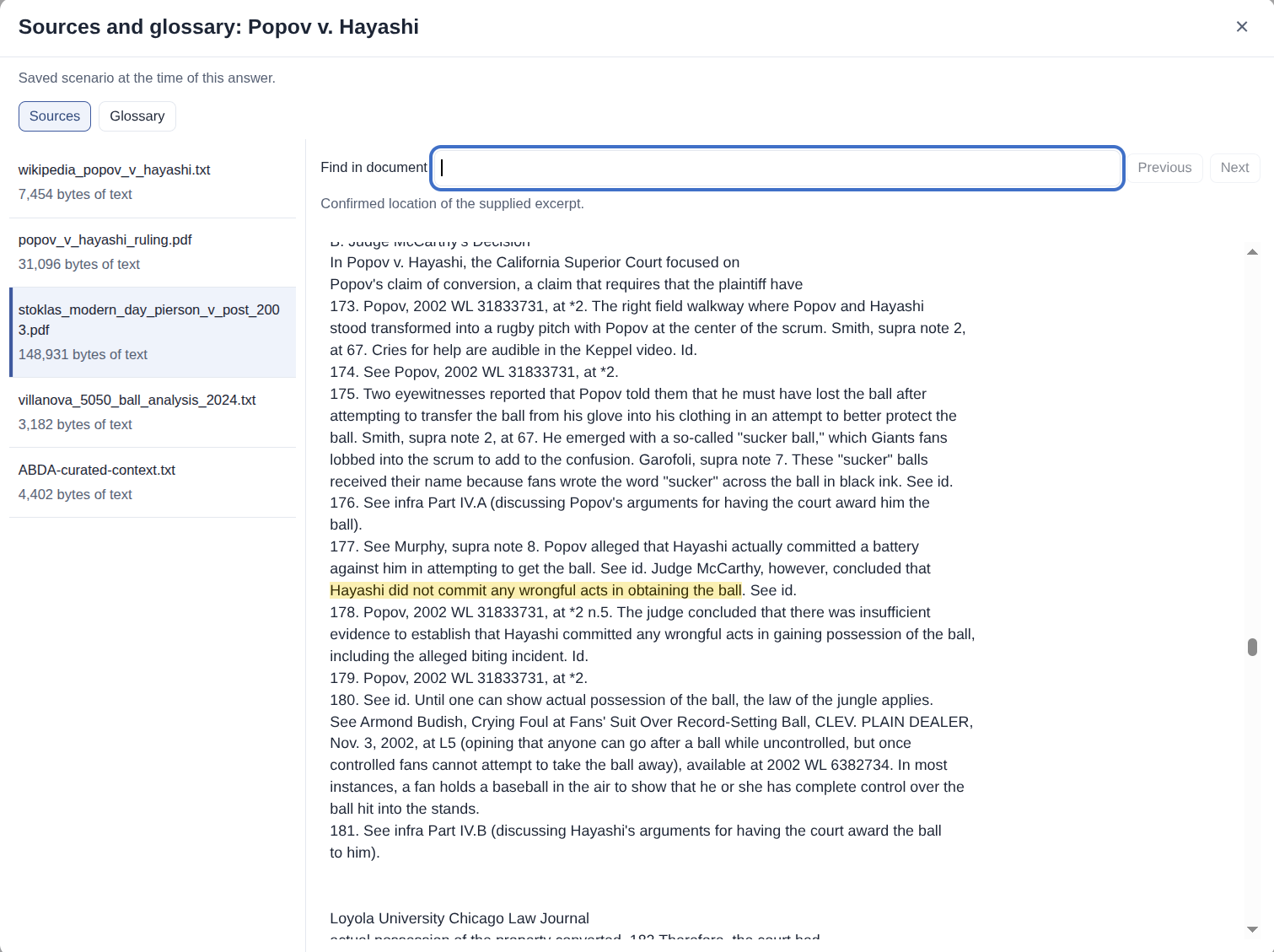}}
  \caption{Creating a scenario in the \ui{Guided} view, with its labels
    computed by \abda in the preview (left), and the sources reader
    showing a quotation verified against the saved document (right).}
  \label{fig:editor}
\end{figure}

\paragraph{Scenario authoring and exchange}
Users are no longer limited to the scenarios we ship.  A unified
editor (\figref{fig:editor}, left) lets them build their own: in the
\ui{Guided} view, statements are written in plain English, typed as
facts, assumptions, or claims, and connected by \kw{if}/\kw{then}
rules marked as strict or defeasible, while the \ui{Rule text} view
shows and edits the same knowledge base in \aspicm syntax.  Statement descriptions double as the glossary,
symbols can be renamed with every reference rewritten consistently,
and reference documents (text, Markdown, or text-based PDF) can be
attached to give the chat something to cite, without becoming logical
facts.  \ui{Check \& save} validates the draft and saves it as a
private scenario, \ui{Preview} shows the computed \abda results
first, and neither needs a model call.  Scenarios can be imported
from YAML or JSON files or from \aspicm rule text with a glossary,
and exported as one self-contained JSON file that carries the full
reference text and opens on any installation.

\paragraph{Private scenarios, sharing, and community scenarios}
Signed-in users save their work as private scenarios, which then
serve as the baseline for further exploration and questions; saves
are version-checked, so two browser tabs cannot silently overwrite
each other.  A scenario can be shared through a revocable read-only
link: recipients can inspect and explain it but not modify it, and
may save a private copy to continue on their own.  A scenario that
others might learn from can be suggested to the community; after the
author's explicit consent and a curator's review, it is listed next
to the built-in scenarios for every visitor.

\paragraph{Chat and exploration}
The chat has grown into a workspace for exploring a scenario.  Items
added with the sparkle icon become removable references inside an
editable question, so a user can point at a rule or a conclusion and
ask about it in their own words.  Conversations are saved and can be
exported; any earlier question can be edited into a new branch or
retried, and each question keeps the scenario state it was asked about.  Evidence cards open the cited passage in a reader
for the scenario's documents and glossary (\figref{fig:editor},
right).  On the formal side, a \ui{Derivation} view beside
\ui{Explain} shows any single argument with its premises,
sub-arguments, rules, label, and attacks, linking each
natural-language item to its exact \aspicm counterpart.

\paragraph{Agent access}
\abdanl also exposes a Model Context Protocol
(MCP)\footnote{\href{https://modelcontextprotocol.io}{\texttt{modelcontextprotocol.io}}}
server, so that AI coding agents such as Codex and Claude Code can
operate the tool under a personal, revocable token with explicit
scopes.  Read tools list and open built-in, community, and private
scenarios together with their grounded outcomes; write tools create
private scenarios and apply version-checked edits, after which the
engine recomputes; and two metered tools ask grounded questions and
request proposals.  A proposal returned to an agent is never applied
by the server: the agent must review and apply it explicitly, subject
to the same checks as in the browser.  Reading and editing consume no
model credit, so an agent's own model can drive an explore, edit, and verify loop
against \abda.

\section{Demonstration and Availability}
\label{sec:demo}

The source code of \abdanl is available at
\href{https://github.com/idaks/ABDA-NL}{\nolinkurl{github.com/idaks/ABDA-NL}}
and the hosted service at
\href{https://demo.abda-nl.org}{\nolinkurl{demo.abda-nl.org}}; the
system was demonstrated at COMMA 2026.

\vspace{12pt}

\noindent
\textbf{Acknowledgments.} Supported in part by a Cardiff--DPI seed
grant (XAI-CA project).  This work used Microsoft Azure and Google
Cloud through the CloudBank project \cite{CloudBank21}, which is
supported by National Science Foundation grant \#1925001.  This work
also used CloudBank at UC San Diego through allocation CIS260773 from
the Advanced Cyberinfrastructure Coordination Ecosystem: Services \&
Support (ACCESS) program \cite{ACCESS23}, which is supported by U.S.
National Science Foundation grants \#2138259, \#2138286, \#2138307,
\#2137603, and \#2138296.  \abdanl embeds the \abda engine
\cite{CU20}, whose implementation by S\"oren Uebis is gratefully
acknowledged.

\clearpage

\bibliographystyle{vancouver}
\bibliography{extended}

\end{document}